\documentclass[conference]{IEEEtran}
\usepackage{geometry}
\usepackage{tikz}
\usepackage{circledsteps}
\usepackage{multirow}
\usepackage{multicol}
\usepackage{subfigure}
\usepackage{tikz}
\usepackage{circledsteps}
\usepackage{multirow}
\usepackage{multicol}
\usepackage{makecell}
\usepackage{subfigure}
\usepackage[normalem]{ulem}
\usepackage{cite}
\usepackage{amsmath,amssymb,amsfonts}
\usepackage{algorithmic}
\usepackage{graphicx}
\usepackage{textcomp}
\usepackage{xcolor}
\usepackage{soul}
\begin{document}

\title{\Large\bf Fast Cell Library Characterization for Design Technology Co-Optimization Based on Graph Neural Networks\\~\\
	}

\author{
    \IEEEauthorblockN{Tianliang Ma, Guangxi Fan, Zhihui Deng, Xuguang Sun, Kainlu Low, Leilai Shao*}
    \IEEEauthorblockA{Shanghai Jiao Tong University, Shanghai, China}
    \IEEEauthorblockA{*Corresponding author: Leilai Shao (leilaishao@sjtu.edu.cn)}
}

\maketitle

\makeatletter
\def\ps@IEEEtitlepagestyle{%
  \def\@oddfoot{\mycopyrightnotice}%
  \def\@evenfoot{}%
}
\makeatother
\def\mycopyrightnotice{%
  \begin{minipage}{\textwidth}
    \footnotesize
    ~ \hfill\\~\\
  \end{minipage}
  \gdef\mycopyrightnotice{}
}

{\small\bf Abstract---
Design technology co-optimization (DTCO) plays a critical role in achieving optimal power, performance, and area (PPA) for advanced semiconductor process development. Cell library characterization is essential in DTCO flow, but traditional methods are time-consuming and costly. To overcome these challenges, we propose a graph neural network (GNN)-based machine learning model for rapid and accurate cell library characterization. Our model incorporates cell structures and demonstrates high prediction accuracy across various process-voltage-temperature (PVT) corners and technology parameters. Validation with 512 unseen technology corners and over one million test data points shows accurate predictions of delay, power, and input pin capacitance for 33 types of cells, with a mean absolute percentage error (MAPE) $\le$ 0.95\% and a speed-up of 100X compared with SPICE simulations. Additionally, we investigate system-level metrics such as worst negative slack (WNS), leakage power, and dynamic power using predictions obtained from the GNN-based model on unseen corners. Our model achieves precise predictions, with absolute error $\le$3.0 ps for WNS, percentage errors $\le$0.60\% for leakage power, and $\le$0.99\% for dynamic power, when compared to golden reference. With the developed model, we further proposed a fine-grained drive strength interpolation methodology to enhance PPA for small-to-medium-scale designs, resulting in an approximate 1-3\% improvement.}




\section{Introduction}
With the slowdown of Moore's Law, design technology co-optimization (DTCO) has become crucial for continuing the rapid scaling of silicon transistors. DTCO involves jointly considering manufacturing processes and system performance to guide developments in advanced technology nodes and evaluate emerging materials/devices. Unlike traditional technology development, where process optimization and system performance evaluations are separate stages, DTCO integrates system performance and reliability considerations early in the process and device optimization \cite{dtco_describtion,iccad:germany,dtco_highp}. 
\begin{figure}[htb]
\centering
\includegraphics[width=0.8\linewidth]{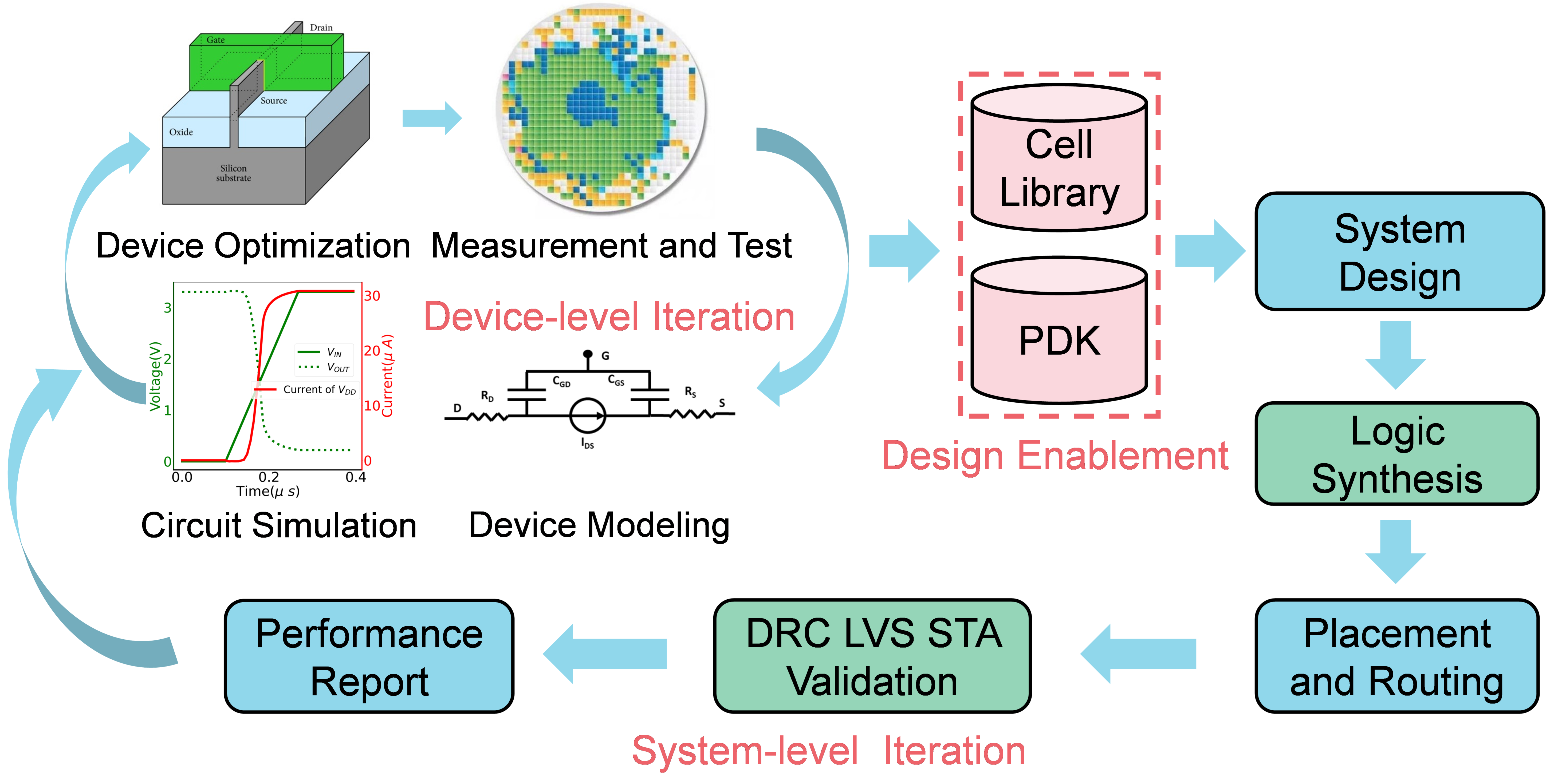}
\caption{Cell library serves as the bridge between process development and performance evaluations of the system-level DTCO iterations.}
\label{fig:intro}
\end{figure}
In comparison to conventional staged optimization flows, DTCO offers greater potential for improving overall system performance and reliability by simultaneously exploring materials, device structures, manufacturing processes, and system design choices. Achieving this requires seamless collaboration among process engineers, technology computer-aided design (TCAD) engineers, and circuit designers \cite{dtco_backend, dtco_device}.

Fig. \ref{fig:intro} illustrates the DTCO flow, which can be categorized into two main optimization scopes: device-level iterations and system-level iterations. The cell library plays a critical role as it serves as the link between detailed process information and the performance/reliability of digital systems, as shown in Fig. \ref{fig:intro}. The cell library not only needs to provide accurate predictions for circuit-level and system-level power, performance, and area (PPA) but also needs to evaluate the reliability of a specific process or emerging materials, such as aging degradation and thermal stability. 

A cell library comprises timing, power, and layout information for various standard cells based on given timing arcs, output loads, and input transitions. The library can vary significantly for different process-voltage-temperature (PVT) corners. Thus, during system-level DTCO iterations, especially for the process development of emerging materials, it becomes necessary to characterize the cell library over a wide range of voltages, temperatures, and threshold voltages whenever materials or device structures are updated. Additionally, when evaluating the aging-induced reliability of a new process/device, such as hot carriers injections (HCI) and negative bias temperature instability (NBTI), the cell library needs to be characterized across a wide range of aging-induced threshold voltage shifts ($\Delta V_{th}$) \cite{temperature effects of circuits,addam:19, germany_group:aging}. Traditionally, cell library characterization is performed through extensive SPICE simulations. However, this approach becomes increasingly expensive and time-consuming as the number of analysis corners grows dramatically for advanced technology nodes and emerging technologies \cite{iccad:canada, iccad:germany}. Thus, there is a strong need for an efficient and cost-effective method of cell library characterization to enable fast and accurate system-level DTCO iterations. 

In recent years, machine learning (ML) methods have been applied to address challenging design automation problems, such as congestion estimations, global placement and detailed routing \cite{synthesis,congestion estimation, placement, detail routing}. For cell library characterization, several studies have aimed to improve computational efficiency by training linear regression and multilayer perceptron (MLP) models \cite{iccad:canada, iccad:germany}. However, these methods either require training one model for each specific technology corner or one model for each specific cell in the library, resulting in the need for dozens or even hundreds of models to be trained and maintained in practice. From a machine learning perspective, this may be due to the limited generality of linear regression and MLP models in processing graph-structured data. Therefore, leveraging the strong abstraction capabilities of graph neural networks (GNNs) holds promise for addressing this issue.  

In this paper, we propose a GNN-based machine learning model for rapid and accurate cell library characterization across a wide range of PVT corners. To the best of our knowledge, this study represents the first comprehensive exploration of GNN for cell library characterization. The primary contributions of this paper can be summarized as follows:
\begin{itemize}
\item Development of a GNN-based machine learning model: We propose a novel approach that utilizes GNNs to accurately characterize and generate cell libraries. Our model takes into account cell structures and demonstrates remarkable capabilities in various prediction tasks for standard cells across diverse PVT corners and technology parameters.
\item Comprehensive validation and performance evaluation: We validate the developed GNN-based model with 512 unseen technology corners, covering mature silicon and emerging carbon-nanotube technologies and a total of over one million test data. The results show that our model achieves accurate predictions of delay, power, and input pin capacitance for 33 types of cells, with a mean absolute percentage error (MAPE) of $\le$ 0.95\% and a speed-up of 100X compared with SPICE simulations.
\item Investigation of system-level metrics: We investigate the system-level worst negative slack (WNS), leakage power and dynamic power based on the predictions obtained from the GNN-based model on an unseen corner. Our model achieves accurate predictions with errors of $\le$ 3.0 ps for WNS, $\le$ 0.60\% percentage error for leakage power and $\le$ 0.99\% percentage error for dynamic power, as compared to the golden reference. These findings demonstrate the model's capability for rapid system evaluations.
\item Introduction of fine-grained drive strength interpolation methodology: We introduce a practical methodology to interpolate fine-grained drive strength into the original standard cell library. Through this approach, we achieve approximately a 1-3\% improvement in PPA for small-to-medium-scale designs.
\end{itemize}

The remaining of this paper is organized as follows: Section \ref{Our Proposed machine learning-based approach} provides a detailed description of our machine learning method. Section \ref{Experimental Results and Analysis} presents the validations of our model on cell library characterizations, system-level evaluations and drive strength interpolation results. Section \ref{Conclusions} draws the conclusion.

\section{Graph Neural Network based Methodology}
\label{Our Proposed machine learning-based approach}
\begin{figure*}[tb]
\centering
\includegraphics[width=1\linewidth]{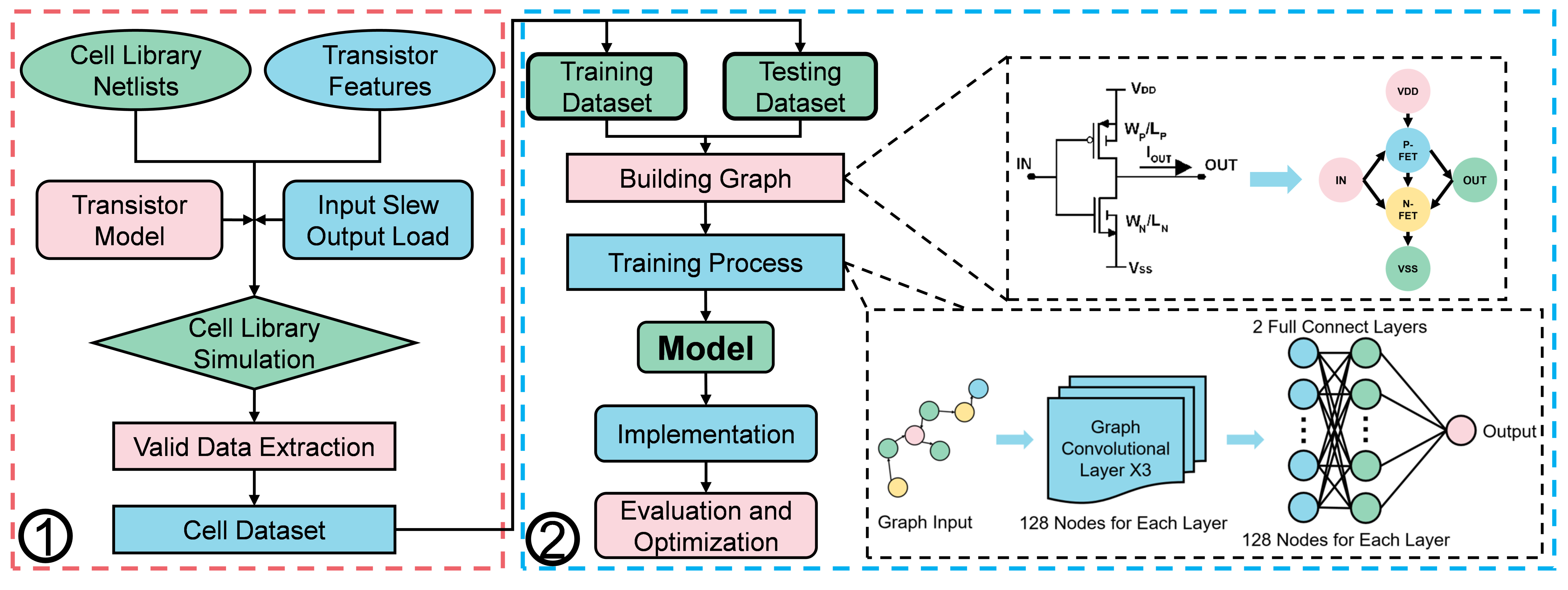}
\caption{The structure of our fast cell library characterization method: from dataset generation to design performance evaluation. \textcircled{1} represents the cell characterization process for dataset generation; \textcircled{2} describes the model training and implementation process. }
\label{fig:whole_flow}
\end{figure*}

Our fast cell characterization method, as shown in Fig. \ref{fig:whole_flow}, consists of two main parts. In the first part, we employ transistor-level SPICE simulation to generate large cell datasets including delay, leakage power, capacitance (maximum capacitance of each input pin), flip power (dynamic power generated when both input and output pins are flipped), and non-flip power (dynamic power generated when only the input pins are flipped, while the state of output pin remains unchanged) under different process-voltage-temperature (PVT) corners and technology settings. Once the data is extracted, it is transformed into a directed graph format using PyTorch Geometric \cite{Pytorch_geo_intro}. The second part involves building a machine learning model for the generated graph data, training, validation and implementation.


In order to showcase the effectiveness of our method, we develop a comprehensive cell library that contains 33 types of combinational cells. The library details are summarized in Table \ref{tab:cells}. During cell library characterization, gate delay and power consumption are captured based on the input transition time of the cell and the effective capacitive load of the output.
\begin{table}[htbp] 
    \centering
    \caption{Standard Cell Library for dataset generation}
    \begin{tabular}{c c}
    \hline
    Logic Cells  & Drive Strengths \\
    \hline
    AND2, OR2, NAND2, NOR2 & X1,X2,X4\\
    AND3, OR3, NAND3, NOR3 & X1 \\
    AOI21, OAI21 & X1 \\
    MX2, XOR2, XNOR2 & X1, X2 \\
    INV &X1,X2,X4,X8,X16\\
    BUF &X2,X4,X8,X16\\
    \hline
    \end{tabular}
    \label{tab:cells}
\end{table}


\subsection{Cell Graph Building}
The primary focus of this paper is on combinational cells, as sequential cells exhibit greater complexity in terms of structure and timing characteristics. As illustrated in Fig. \ref{fig:whole_flow}, each cell can be transformed into a directed graph. Our graph consists of five types of nodes, distinguished using three bits: $IN (001)$, $OUT (010)$, $FET (011)$, $V_{DD} (100)$, and $V_{SS} (101)$. The directions between nodes are defined as follows: for the $IN$ and $V_{DD}$ nodes, all edges originate from that node and connect to other nodes; for the $OUT$ and $V_{SS}$ nodes, all edges lead to that node from other nodes. In the case of the $FET$ node, edges at the gate and source terminals originate from other nodes, while the edge at the drain terminal connects to other nodes. To differentiate between N- and P-type FETs, we utilize an additional bit, with "1" representing P-type and "-1" representing N-type. Moreover, we include the transistor width (Width) from the cell netlist as part of the node feature vector to distinguish cells with varying drive strengths.

In the context of silicon technology, our focus lies in analyzing the variation of supply voltage ($V_{DD}$), threshold voltage ($V_{th}$), and transistor temperature (Temperature) across various unseen PVT corners. The definitions of node features are presented in Table \ref{tab:node_features_45}. Here, "Current\_state" and "Next\_state" are two features related to the input pin state, with "1" denoting a high level and "-1" representing a low level, respectively. "Input\_slew" denotes the transition time of input signal, "Output\_load" denotes capacitive load on cell output pins.
\begin{table}[htbp] 
    \centering
    \caption{Node feature vector definition for silicon technology }
    \resizebox{\linewidth}{!}{
    \begin{tabular}{ccccccc}
    \hline
    &  IN& OUT& N-FET& P-FET& $V_{DD}$& $V_{SS}$\\
    \hline
     Bit0&       0& 0& 0& 0& 1& 1\\     
     Bit1&       0& 1& 1& 1& 0& 0\\
     Bit2&       1& 0& 1& 1& 0& 1\\
     Bit3&       0& 0& -1& 1& 0& 0\\
     Bit4&       0& 0& 0& 0& $V_{DD}$& 0\\
     Bit5&       0& 0& Width& Width& 0& 0\\
     Bit6&       0& 0& Temperature& Temperature& 0& 0\\
     Bit7&       0& 0& $V_{th}$& $V_{th}$& 0& 0\\
     Bit8&       Input\_slew& 0& 0& 0& 0& 0\\
     Bit9&       0& Output\_load& 0& 0& 0& 0\\
     Bit10&      Current\_state& 0& 0& 0& 0& 0\\
     Bit11&      Next\_state& 0& 0& 0& 0& 0\\
    \hline
    \end{tabular}
    \label{tab:node_features_45}
    }
\end{table}

Specifically, when training the model for leakage power prediction, since it is unrelated to input pin flips and solely depends on the current state, the node feature definition is modified to [$Type_{0}, Type_{1}, Type_{2}, Polar, V_{DD}, Width$, $Temperature, V_{th}, Current\_state$].
When training the model for capacitance prediction, as our goal is to estimate the capacitance of each input pin, the node feature definition is modified to [$Type_{0}, Type_{1}, Type_{2}, Polar, V_{DD}, Width$, $Temperature, V_{th}, Pin\_is\_chosen$]. For "Pin\_is\_chosen", a value of "1" indicates that capacitance prediction is performed on this input pin, while a value of "0" indicates that this pin is not selected for capacitance prediction within the graph.

In our study of emerging technology, we utilize an unified flexible compact model \cite{leilaishao:flex_model} and specifically focus on analyzing the variation of supply voltage ($V_{DD}$), threshold voltage ($V_{th}$), and gate unit capacitance ($C_{ox}$). This emphasis is due to the fact that the flexible compact model we employ does not account for temperature variation. Additionally, $C_{ox}$ is a critical parameter that significantly influences the performance of flexible devices. The feature definition for each node remains consistent with the silicon technology case, except for one modification: $Temperature$ feature is replaced with the $C_{ox}$ feature. 

\subsection{Machine Learning Model Structure}
Graph neural networks (GNNs) have gained significant attention in the field of design automation due to their inherent ability to represent and process circuit data in the form of graphs. In this study, we adopt a classic GNN method known as Graph Convolutional Network (GCN) to establish our framework. Considering the relatively low complexity of a cell graph and the optimal performance typically observed with two or three layers in GCN \cite{GCN:layer_num}, we choose three-layer GCN to build our model. To enhance the accuracy of predictions, we added two additional fully connected layers after the three GCN layers. Consequently, our overall model is composed of a five-layer deep neural network, with ReLU as the activation function for each layer which consists of 128 neurons.




\section{Experimental Results and Analysis}
\label{Experimental Results and Analysis}
In this section, thorough evaluations were conducted to assess the performance of the developed model for cell prediction across diverse unseen PVT corners and technology parameters. The focus was placed on two representative technologies: 45nm silicon technology and 600nm carbon-nanotube (CNT) flexible technology. The model's capability to predict timing and power at the system level on unseen PVT corners was explored. Furthermore, case studies were performed to examine the effectiveness of the developed cell library prediction model in fine-grained drive strength interpolation, with the goal of enhancing power, performance, and area (PPA) metrics for chip design.
\subsection{Experiment setup}
Our cell characterization model was implemented using the PyTorch machine learning framework on a Linux workstation equipped with an Intel Xeon Gold 6230 (2.1GHz) CPU, 128GB of RAM and a NVIDIA Quadro RTX5000 graphic card.

For the 45nm silicon technology, we employed the Cadence GPDK045 BSIM model \cite{gpdk045_descrip} for both NMOS and PMOS transistors. The variable ranges used in our model were consistent with those employed in a previous study \cite{iccad:canada}. These ranges include a temperature range of $T:[20:120]^{o}C$, threshold voltage range of $V_{th}:[0.1:0.5]V$ for NMOS transistors and $[-0.5:0.1]V$ for PMOS transistors, an input transition range of $[5:950]ps$, an output capacitive load range of $[0.25:25]fF$, and a supply voltage range of $V_{DD}:[0.9:1.1]V$.

For the flexible technology, we determined the variable ranges based on crucial parameters from a state-of-the-art carbon nanotube model \cite{CNT_para}. Specifically, the threshold voltage range was set to $V_{th}:[0.3:1.1]V$ for NMOS transistors and $V_{th}:[-1.1:-0.3]V$ for PMOS transistors. The gate unit capacitance range was established as $C_{ox}:[50:130]nF/cm^{2}$. The input transition range was defined as $[1:100]ns$, and the output capacitive load range was selected as $[0.1:300]fF$. The supply voltage range was set to $[0.5:2.5]V$.

For the training dataset, we selected five equally spaced values within the variable ranges for each technology, resulting in 125 PVT corners. Using these corners, we generated 125 cell libraries and extracted the valid data related to cell delay, power, and capacitance to construct the training dataset.

\begin{figure}[htbp]
\centering
    \begin{minipage}{.49\linewidth}
    \centering
    \includegraphics[width=1.0\linewidth]{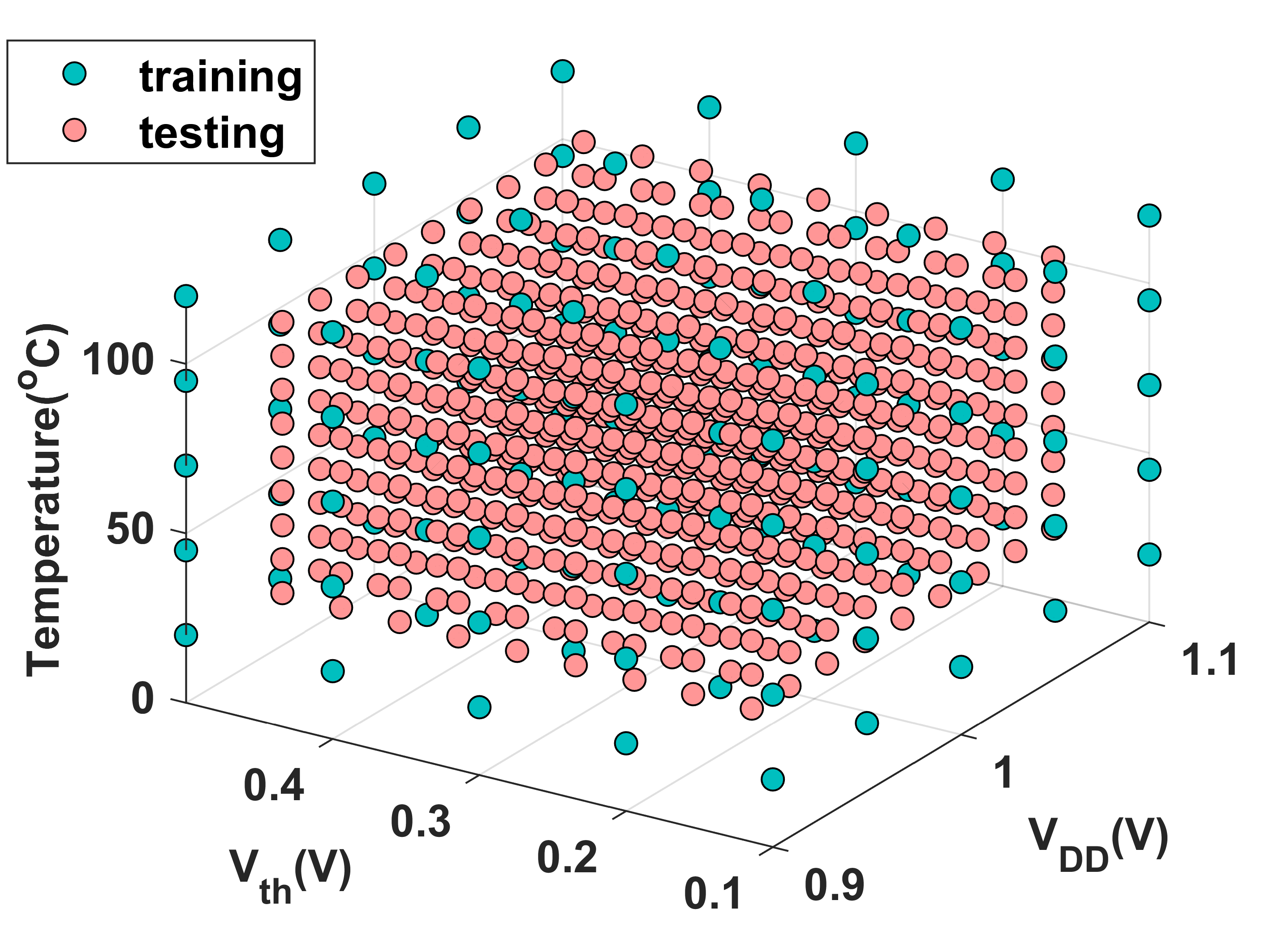}
    \end{minipage}
    \begin{minipage}{.49\linewidth}
    \centering
    \includegraphics[width=1.0\linewidth]{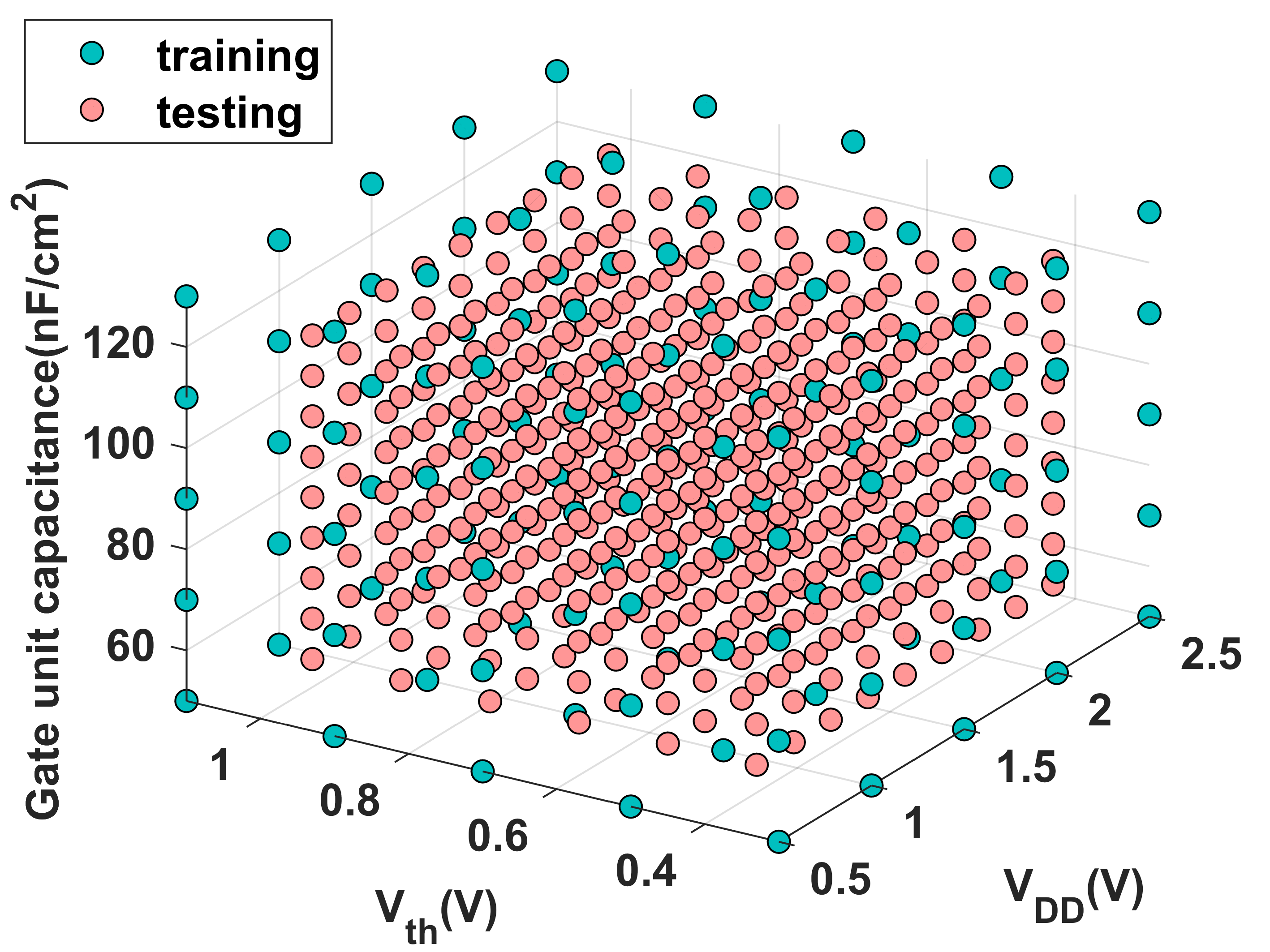}
    \end{minipage}
\caption{Distribution of training and testing corners for 45nm silicon technology (left) and flexible technology (right).}
        \label{fig:my_label}
\end{figure}

For the testing dataset, we chose eight equally spaced values within the variable ranges for each technology, resulting in 512 PVT corners. These corners were used to extract valid data and generate the testing dataset. The distribution of training and testing points is illustrated in Fig. \ref{fig:my_label}, showing a uniform distribution.

Before training the proposed model, several steps were taken to optimize the training process:

Normalization: We use min-max normalization to normalize all input features into range of [0, 1] to accelerate training, as shown in Equation (\ref{eq.min-max-normalization}). 

\begin{equation}\label{eq.min-max-normalization}
x_{\text{norm}} = \frac{x - x_{\text{min}}}{x_{\text{max}} - x_{\text{min}}}
\end{equation}

In the equation, $x$ represents the original value of the input feature, $x_{\text{min}}$ and $x_{\text{max}}$ are the minimum and maximum values of the input feature, respectively, and $x_{\text{norm}}$ is the normalized value of the input feature.

Weight Initialization: We adopted the widely-used He initialization method \cite{He_init} in our model to ensure that the weights are initialized with appropriate values to alleviate problems such as vanishing or exploding gradients during training.

Loss Function: Due to the significant variation in cell performance values across different PVT corners and technology parameters, the Mean Absolute Percentage Error (MAPE) is chosen as the loss function. MAPE, as shown in Equation (\ref{eq.MAPE_function}), measures the percentage difference between the predicted ($\hat{Y_{i}}$) and actual ($Y_{i}$) values. By minimizing MAPE during training, the model is optimized to accurately predict both small and large cell performance values. 

\begin{equation}\label{eq.MAPE_function}
\text{MAPE} = \frac{1}{N}\sum_{i=1}^{N}\left|\frac{\hat{Y_{i}}-Y_{i}}{Y_{i}}\right| \times 100\%
\end{equation}

In the equation, $\hat{Y_{i}}$ represents the predicted value, $Y_{i}$ is the actual value, and $N$ is the total number of samples.

Optimization Algorithm: The state-of-the-art Adam optimization algorithm \cite{Adam_algorithm} is utilized to minimize the loss function during training. We set training batch size to 512, and the model is trained for 5000 epochs to ensure convergence. To enhance training stability and prevent the model from getting stuck in local minima, a dynamic learning rate is employed. The initial learning rate is set to 0.0001 and is halved after every 500 epochs. For one technology, the total training time for five models is less than 16 hours.

\subsection{Accuracy of cell library prediction}
To assess the accuracy of our proposed model, we evaluate its performance by comparing the predicted values with the ground truth values of cell performance on the testing dataset. The evaluation is conducted using the Mean Absolute Percentage Error (MAPE) metric. The MAPE values for both the 45nm silicon technology and the flexible technology are found to be consistently low across all aspects of cell performance, indicating the model's high predictive accuracy. Detailed results can be found in Table \ref{tab:MAPEs_test_gpdk045_flex}.
\begin{table}[htbp] 
    \centering
    \caption{MAPEs of cell library prediction in the whole testing dataset}
    \resizebox{\linewidth}{!}{
    \begin{tabular}{cccc}
    \hline
    & 45nm Silicon& Emerging Flexible& Number of Data Points\\
    \hline
    Delay&  0.47\%& 0.62\%& 654650\\  
    Capacitance& 0.24\%& 0.21\%& 66066\\
    Flip Power& 0.77\%& 0.40\%& 654650\\
    Non-flip Power& 0.95\%& 0.71\%& 339156\\
    Leakage Power& 0.25\%& 0.39\%& 143944\\
    \hline
    \end{tabular}
    \label{tab:MAPEs_test_gpdk045_flex}
    }
\end{table}

To provide a fair comparison with similar state-of-the-art works, we adopt the same evaluation indicators as used in \cite{iccad:canada} and \cite{iccad:germany}. For the 45nm silicon technology, the indicator used is the Root Mean Squared Percentage Error (RMSPE), where lower values indicate better performance. For the emerging flexible technology, the indicator used is the $R^{2}$ score, where values closer to 1 indicate higher accuracy.

\begin{figure}[htbp]
    \centering
    \includegraphics[width=0.9\linewidth]{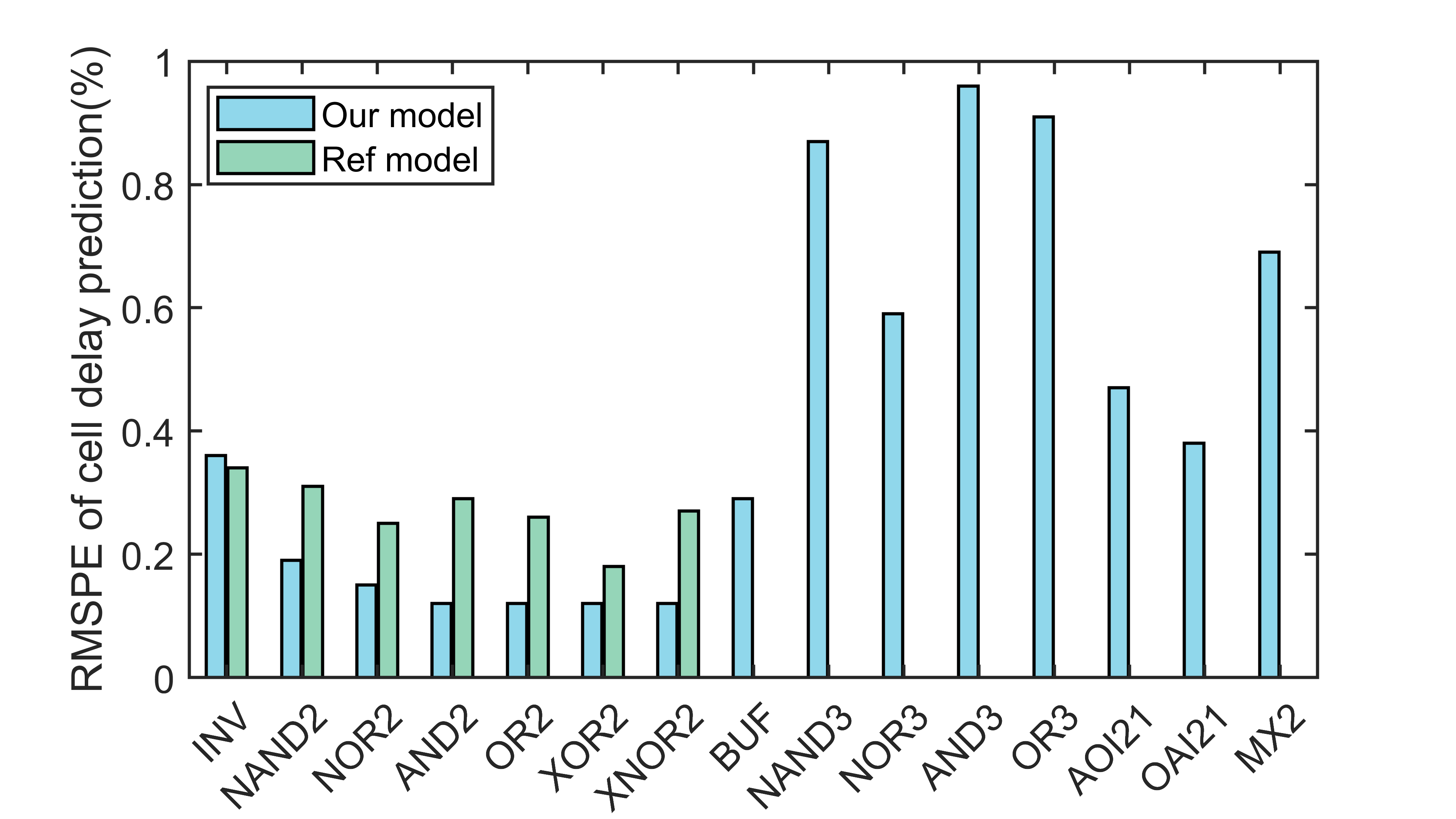}
    \caption{Prediction comparison between our proposed model and reference model in 45nm silicon technology about cell delay.}
    \label{fig:45_rmspe_compare}
\end{figure}

Fig. \ref{fig:45_rmspe_compare} compares the results of our proposed model with similar state-of-the-art feedforward neural network (FFNN) models for cell delay prediction in the 45nm silicon technology. In comparison to the results reported in \cite{iccad:canada} for seven cells, our proposed model achieves lower prediction errors in six out of the seven cells. Furthermore, the RMSPE values for more complex cells in our library are also lower than 1\% with our proposed model.

\begin{figure}
\centering
    \begin{minipage}{.49\linewidth}
    \centering
    \includegraphics[width=1.0\linewidth]{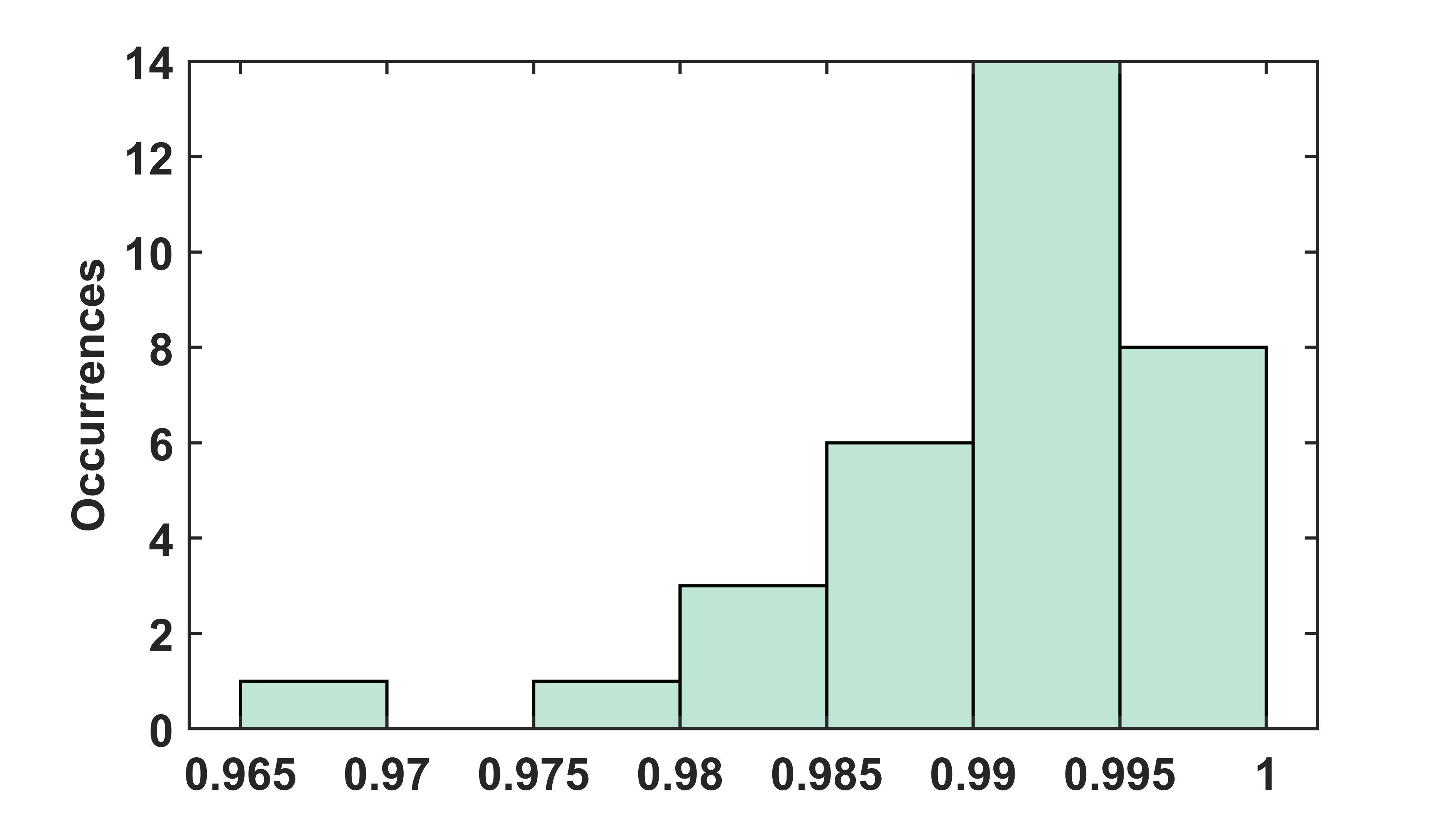}
    \end{minipage}
    \begin{minipage}{.49\linewidth}
    \centering
    \includegraphics[width=1.0\linewidth]{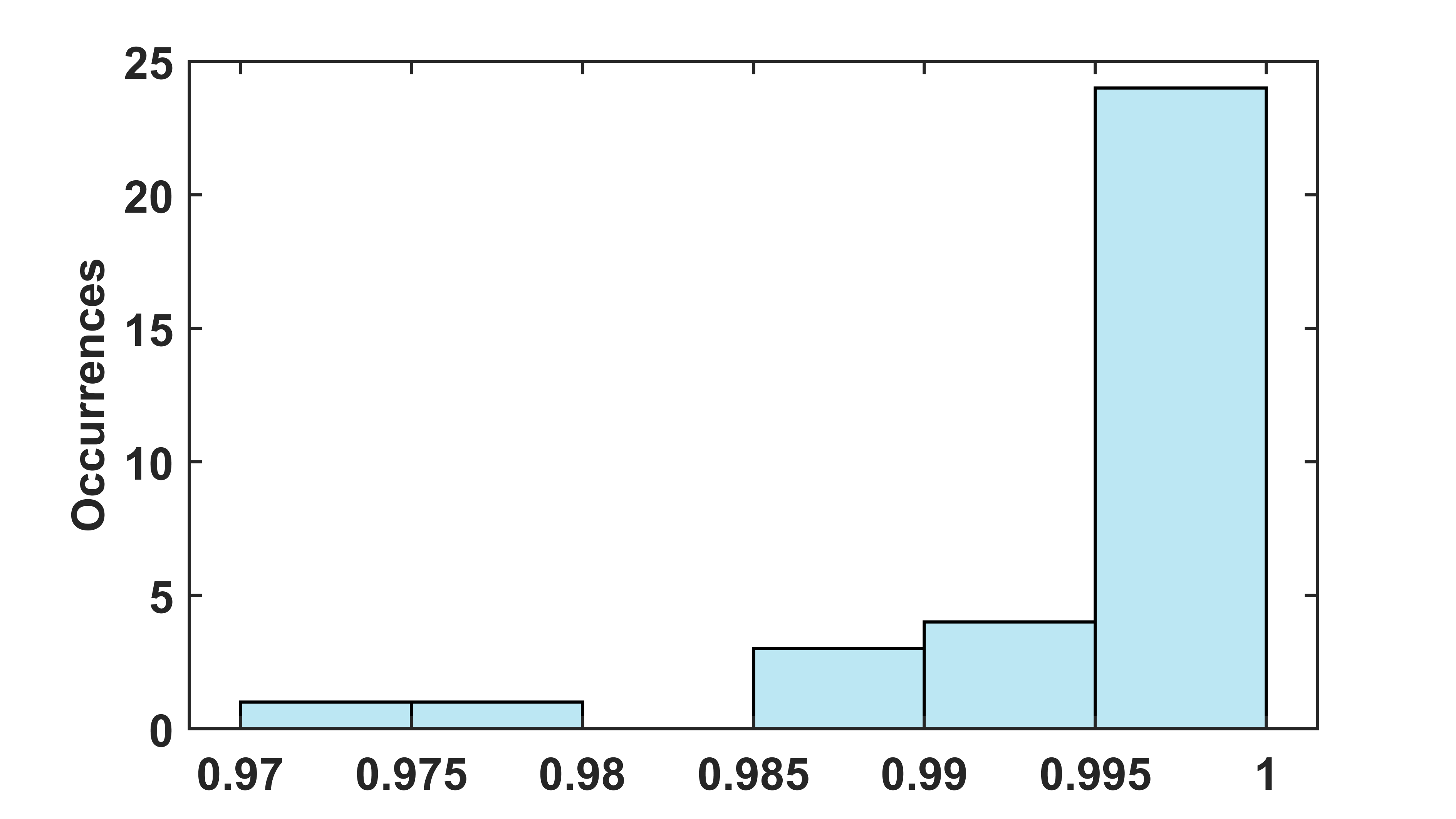}
    \end{minipage}
\caption{Histograms of $R^{2}$ score of capacitance (left) and cell delay prediction (right).}
        \label{fig:r2}
\end{figure}

In the case of the emerging flexible technology, our proposed model demonstrates exceptional performance with average $R^2$ scores of 0.9961 for capacitance and 0.9924 for cell delay. These results surpass those of the linear regression model presented in \cite{iccad:germany}, which achieved average $R^2$ scores of 0.9700 for capacitance and 0.9902 for cell delay. Fig. \ref{fig:r2} showcases histograms of the $R^2$ scores for each cell, revealing that the majority of cells achieve an $R^2$ score above 0.99, with the lowest $R^2$ score surpassing 0.965. 
In summary, our proposed model surpasses state-of-the-art works in terms of accuracy and generalization performance. 

\subsection{Accuracy of prediction on system level}
After evaluating the prediction accuracy of each cell, it is important to test whether our proposed model can be applied to system-level prediction. To accomplish this, we selected an unseen PVT corner for the 45nm silicon technology, characterized by the following parameters: $V_{DD}$: 1.03V, $V_{th}$: 0.48V, and Temperature: 65$^\circ$C. On this corner, we first used commercial synthesis tool to generate circuit netlists for benchmarks based on the "ground truth library" (we added one type of DFF for sequential circuits synthesis) generated by a conventional EDA tool and "predicted library" generated by our proposed model. The system-level results were then compared for ten benchmarks, comprising five small-scale sequential circuits (with fewer than 500 gates) including s298, s386, s526, s820, and s1196, three medium-scale circuits from ISCAS89 \cite{ISCAS89}, and two open-source RISC-V cores, namely Picorv32 \cite{picorv32} and Darkriscv \cite{darkriscv}, with up to 20k gates. For all the ten benchmarks, the clock frequency is set to be 500MHz, which is commonly used for 45nm silicon technology. 

Then, We used the same synthesized circuit netlists and constraint files in both commercial static timing analysis (STA) tool and power analysis tool to compare the system-level results. Since our model temporarily only support combinational cells, properties of DFF are the same as in "ground truth library".

As depicted in Table \ref{tab:dynamic_power_result}, the absolute error for worst negative slack (WNS) was $\le$ 3.0 ps across the ten benchmarks, while the percentage error for leakage power is $\le$0.60\% and for dynamic power is $\le$0.99\%. These results affirm that our proposed model can be utilized to rapidly evaluate system-level performance on unseen PVT corners.
\begin{table}[htbp] 
    \centering
    \caption{Comparison of WNS, leakage and dynamic power prediction on system level for 45nm silicon technology between two libraries}
    \resizebox{\linewidth}{!}{
    \begin{tabular}{cccc}
    \hline
    \makecell{Benchmarks}& \makecell{WNS Error ($ps$)}& \makecell{Leakage Power\\Percentage Error}& \makecell{Dynamic Power\\Percentage Error}\\
    \hline
    s298&  0.3& 0.26\%& 0.07\%\\
    s386&  0.9& 0.60\%& 0.13\%\\
    s526&  1.5& 0.17\%& 0.12\%\\
    s820&  2.9&  0.12\%&  0.20\%\\
    s1196&  1.9& 0.22\%&  0.28\%\\
    s1488&  0.6& 0.53\%& 0.50\%\\
    s1423&  1.2& 0.41\%& 0.45\%\\
    s5378&  0.8& 0.39\%& 0.54\%\\
    Picorv32&  0.7& 0.20\%& 0.62\%\\
    Darkriscv&  0.5& 0.38\%& 0.99\%\\
\hline
    \end{tabular}
    \label{tab:dynamic_power_result}
    }
\end{table}

Furthermore, we calculated the runtime for cell library characterization using our proposed model and compared it to traditional SPICE simulations on EDA tool. For simplicity, we only report total time consumption of the SPICE simulations. The results, as shown in Table \ref{tab:run_time_compare}, clearly demonstrate that our machine learning model achieves an acceleration of 100X and the speed-up could be more significant for complete industry cell libraries with hundreds of standard cells. 
\begin{table}[htbp] 
    \centering
    \caption{Runtime comparison between our model and SPICE simulations}
    \begin{tabular}{ccc}
    \hline
    & Our Model& SPICE Simulations\\
    \hline
    Time of Environment Loading& 8.12s& /\\
    Delay&  1.13s& /\\  
    Capacitance& 0.92s& /\\
    Flip Power& 1.15s& /\\
    Non-flip Power& 0.99s& /\\
    Leakage Power& 1.02s& /\\
    Total Time& 13.33s& 1972s\\
    \hline
    \end{tabular}
    \label{tab:run_time_compare}
\end{table}

Overall, our proposed model exhibits high accuracy and efficiency in predicting both cell performance and system-level performance, making it a promising tool for design technology co-optimization (DTCO).

\subsection{Optimization via fine-grained drive strength interpolation}
In common practice, drive strength is typically defined by the number of times the width of the last-stage transistor in a cell is multiplied. However, the original cell libraries provided by foundries often have coarse drive strength granularity, which limits the ability of EDA tools to find the optimal circuit netlist in terms of area and power during the synthesis process \cite{cell granularity}. With our proposed model, designers now have the ability to generate a "Original+plus" cell library that includes cells with more drive strengths and rapidly evaluate their designs using this new library.

We utilized a state-of-the-art carbon nanotube (CNT) technology \cite{CNT_para} with a supply voltage $V_{DD}$ of 2.5V as the foundation for our "Original+plus" cell library characterization. The newly created library comprises various cell types, including AND2X3, NAND2X3, OR2X3, NOR2X3, INVX3, INVX5, INVX6, INVX7, BUFX3, BUFX5, BUFX6, and BUFX7. To determine the area of each new drive strength cell, we manually designed the layout for each of them. The timing and power characteristics of these cells were estimated using our proposed model for system-level evaluation. The MAPE for all five aspects is consistently below 1.5\% for the majority of the cells compared with SPICE simulations, indicating high accuracy of our proposed model in predicting unseen cell properties.

Next, we compared area and power of the same benchmarks used in the previous subsection. Since the clock frequency of flexible systems is typically lower than that of traditional technology, we set the maximum clock frequency to 30MHz for small-scale benchmarks and 20MHz for medium-scale ones. In each synthesis process, we used the "original" library, the "Original+plus (ground truth)" library which combines the original cell library with the 12 new drive strength cells characterized by SPICE simulations, and the "Original+plus (predicted)" library which combines the original cell library with the 12 new drive strength cells characterized by our proposed model. 


In synthesis process, commercial EDA tools are used and WNS is constrained to be positive, performance of each benchmark is equivalent for same clock frequency. To clearly demonstrate the improvement in PPA metrics with a fine-grained drive strength cell library, we calculated the PPA improvement using Eq. (\ref{eq.PPA_improve}).

\begin{small}
\begin{equation}\label{eq.PPA_improve}
PPA_{impro} = \left(1-\frac{Area_{new}}{Area_{origin}} + 1 -\frac{Power_{new}}{Power_{origin}}\right)\times 100\%
\end{equation}
\end{small}

\begin{figure}
\centering
    \begin{minipage}{.49\linewidth}
    \centering
    \includegraphics[width=1.0\linewidth]{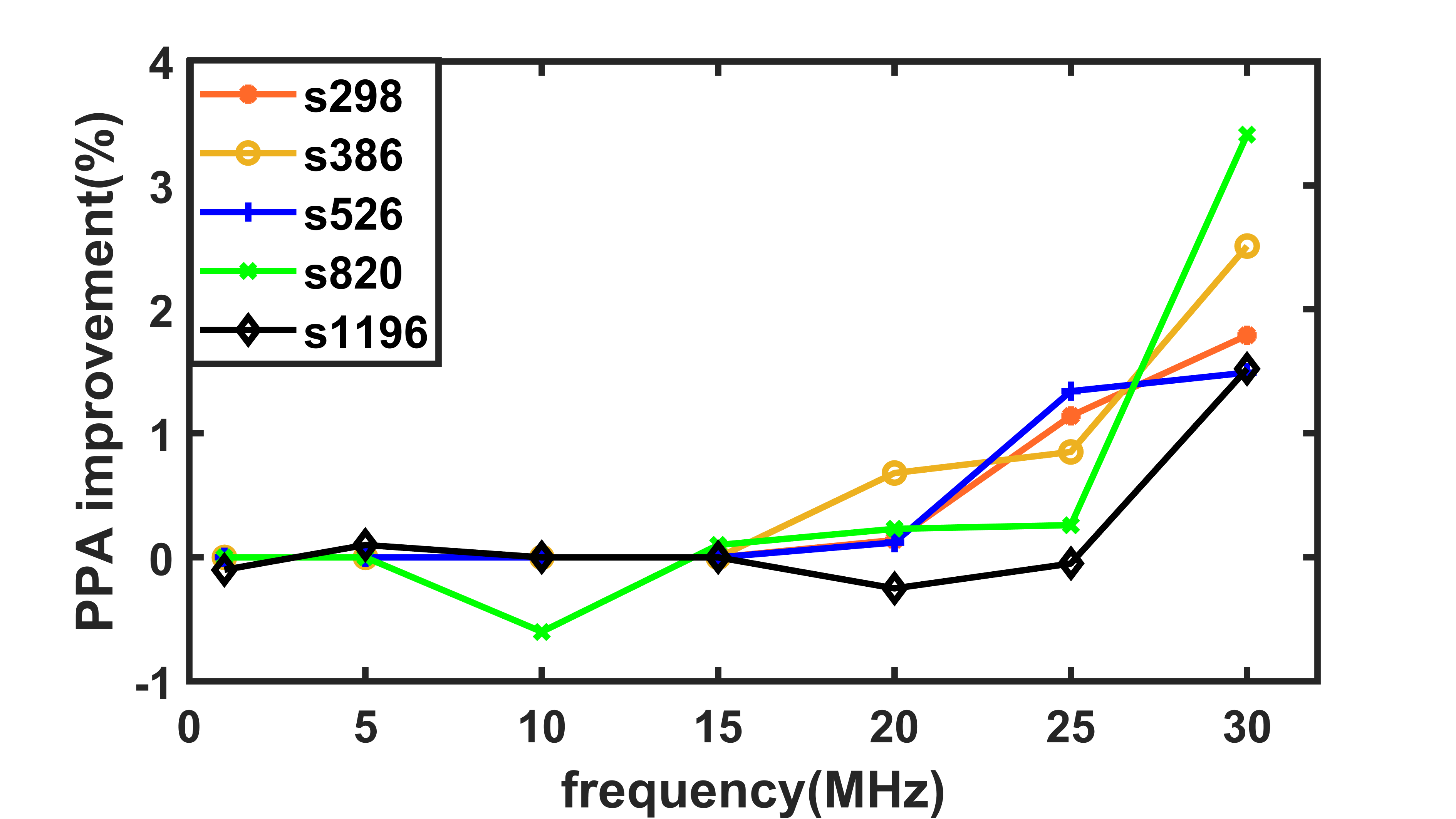}
    \end{minipage}
    \begin{minipage}{.49\linewidth}
    \centering
    \includegraphics[width=1.0\linewidth]{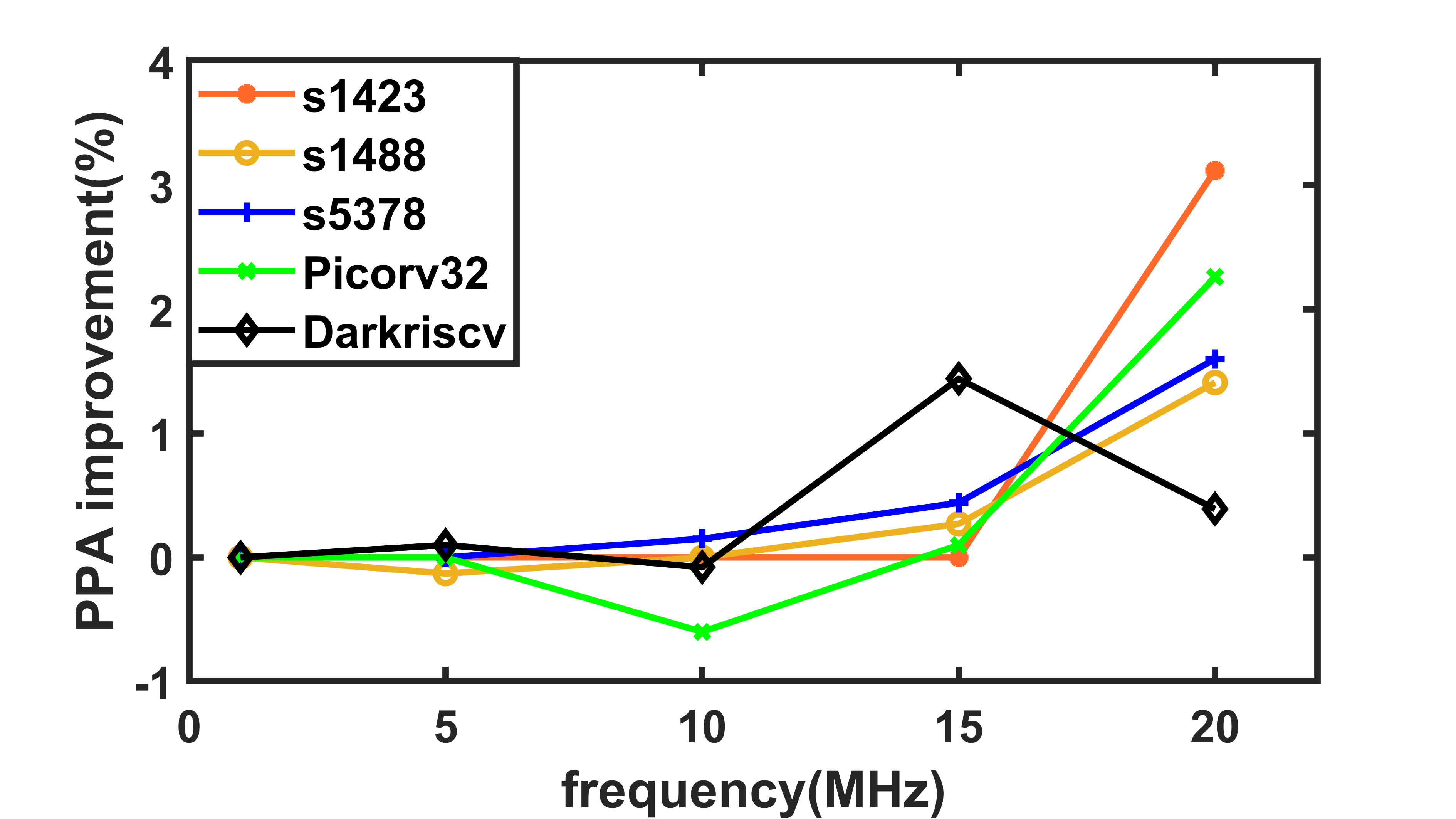}
    \end{minipage}
\caption{PPA improvement of small-scale (left) and medium-scale (right) benchmarks.}
        \label{fig:PPA_improve}
\end{figure}

In Fig. \ref{fig:PPA_improve}, it can be observed that the improvement in power, performance, and area (PPA) is insignificant. However, as the frequency approaches the maximum frequency limit, the PPA improvement becomes significant for most benchmarks, ranging from 1\% to 3.5\%.  These findings highlight the effectiveness of the drive strength interpolation method in meeting high-performance requirements. Our method has shown its potential in cell library optimization for PPA improvement in chip design.


\section{Conclusions}
\label{Conclusions}
In this paper, we present a novel approach for fast and accurate cell library characterization based on graph neural networks (GNNs). Furthermore, we extend the application of our model to the prediction of system level metrics and introduce a fine-grained drive strength interpolation method based on proposed model to enhance the power, performance, and area (PPA) metrics of small-to-medium-scale designs. In the future, we will try to improve the prediction accuracy for extrapolation or out-of-distribution corners and apply it to sequential and more complex cells at advanced nodes for design technology co-optimization (DTCO).





\section*{\sc Acknowledgments}
This material is based upon work supported, in part, by grant from Shanghai Sailing Program (No. 22YF1420100). 



\end{document}